\documentclass{article}

\usepackage{arxiv}

\usepackage[utf8]{inputenc} 
\usepackage[T1]{fontenc}    
\usepackage{hyperref}       
\usepackage{url}            
\usepackage{booktabs}       
\usepackage{amsfonts}       
\usepackage{nicefrac}       
\usepackage{microtype}      
\usepackage{lipsum}
\usepackage{svg}
\usepackage{graphicx}
\graphicspath{ {./images/} }
\usepackage{amsmath} 
\usepackage{pdflscape}
\usepackage{xcolor}
\usepackage[normalem]{ulem}
\usepackage{comment}
 \usepackage{longtable}
\usepackage{tabularx}
\usepackage{caption} 
\usepackage{subcaption}
\usepackage{longtable}
\usepackage{booktabs}
\usepackage[utf8]{inputenc}

\title{ PV-faultNet: Optimized CNN Architecture to detect defects resulting efficient PV production}

\author{
  \textbf{Eiffat E Zaman}\textsuperscript{*} and \textbf{Rahima Khanam}\\[1ex] 
  \begin{minipage}[t]{0.90\textwidth}
    \scriptsize Department of Computer Science, Huddersfield University, Queensgate, Huddersfield HD1 3DH, UK; \\
    \textsuperscript{*}Correspondence: rahima.khanam@hud.ac.uk;
  \end{minipage}
}
\begin{document}
\maketitle
\begin{abstract}The global shift towards renewable energy has pushed PV cell manufacturing as a pivotal point as they are the fundamental building block of green energy. However, the manufacturing process is complex enough to lose its purpose due to probable defects experienced during the time impacting the overall efficiency. However, at the moment, manual inspection is being conducted to detect the defects that can cause bias, leading to time and cost inefficiency. Even if automated solutions have also been proposed, most of them are resource-intensive, proving ineffective in production environments. In that context, this study presents PV-faultNet, a lightweight Convolutional Neural Network (CNN) architecture optimized for efficient and real-time defect detection in photovoltaic (PV) cells, designed to be deployable on resource-limited production devices. Addressing computational challenges in industrial PV manufacturing environments, the model includes only 2.92 million parameters, significantly reducing processing demands without sacrificing accuracy. Comprehensive data augmentation techniques were implemented to tackle data scarcity, thus enhancing model generalization and maintaining a balance between precision and recall. The proposed model achieved high performance with 91\% precision, 89\% recall, and a 90\% F1 score, demonstrating its effectiveness for scalable quality control in PV production.
\end{abstract}


\section{Introduction}
In the context of pursuing sustainable energy, solar power has been placed as the leading technology in renewable energy generation. Being abundant, solar energy plays a vital role in mitigating climate change while curbing the dependence on fossil fuels. According to Blakers \cite{weforum_solar}, the total global Solar photovoltaic (PV) capacity has grown rapidly, with an annual increase of 22\% and projections suggest it will reach 6 terawatts by 2031. Additionally, the global solar market is expected to reach USD 223.3 billion by 2026, driven by government incentives and decreasing PV module prices. However, ensuring the efficiency and quality of PV cells is essential for the continued sector.

Maintaining the efficiency and integrity of PV (Photovoltaic) cells which are the bedrock units \cite{PlaneteEnergies2024} that transform sunlight into electricity through the PV effect—is \cite{jungbluth2009photovoltaics} pivotal for the sustained expansion of the sector \cite{EIA2024} as they produce the highest power in their ideal state \cite{pavlik2023analysis}. Even minor defects like cracks, contamination, or misalignment can significantly affect the efficiency of these cells, thereby hampering the overall performance of solar panels and putting them at risk in terms of the economic feasibility of solar projects. Research indicates that manufacturing anomalies can slash efficiency by 10–15\%, which, when magnified across expansive solar farms, results in considerable energy losses and financial burdens on the industry \cite{kontges2014performance}.

The production of PV cells is complex, involving a series of stages each susceptible to defect introduction, such as wafer slicing, cell doping, and coating. Traditionally, manual inspections during production have been the norm, but this approach is laborious, error-prone, and unsuitable for high-volume production lines \cite{Marsh2023}. Consequently, with the demand for solar panels on an upward trajectory, there is a pressing need for automated, scalable, and precise defect detection systems. Nonetheless, challenges persist, including limitations in data availability, real-time monitoring, measurement accuracy, computational efficiency, and dataset distribution, all of which impact the effectiveness of defect detection in PV cell production \cite{hijjawi2023review}.

In this context, Computer Vision \cite{szeliski2022computer}, especially Convolutional Neural Networks (CNNs) \cite{li2021survey}, has shown great promise in automating defect detection in PV cells \cite{lin2022development, deitsch2019automatic, al2024classification, rahman2021cnn, khanam2024comprehensive}
 . CNNs have established a distinctive identity in image recognition tasks \cite{liu2017review}, making them particularly suited for pinpointing defects by identifying subtle patterns and irregularities that may get overlooked during human inspections. Yet, deploying CNN models in production facilities is not without its hurdles, particularly for resource-limited edge devices that are far from the course in industrial settings. These devices often lack the ability required to process large CNN models in real time.

The development of streamlined, lightweight CNN models tailored for the detection of defective PV cells would bring forth significant opportunities by reducing the computational demand while preserving detection accuracy, paving the way for real-time inspection on edge devices, and ensuring scalable and effective quality control in the manufacture of PV cells.

\section{Literature Review}
After reviewing the current literature on automated PV cell quality inspection, it is evident that researchers are actively investigating the application of deep learning, specifically computer vision techniques, for PV defect identification.

For instance, Acharya et al. \cite{acharya2021deep} utilized a Deep Siamese CNN to classify different types of cracks in solar cells. This architecture, known for comparing pairs of images to identify discrepancies indicating defects, achieved an average multi-class accuracy of 74.75\% and an AUC of 0.90. However, despite its robustness, the model also can be complex and resource-intensive, requiring significant computational resources and expertise.

In contrast, Akram et al.\cite{akram2019cnn} developed a lightweight CNN model that operates efficiently on standard CPU systems, achieving real-time processing speeds of 8.07ms per image and maintaining a high accuracy rate of 93.02\%. Through iterative testing and fine-tuning, they settled on a six-layer CNN architecture comprising four convolutional layers and two fully connected layers. The model was trained for 150 epochs with a batch size of 64. To combat data scarcity—a common challenge in deep learning—they applied data augmentation techniques such as rotation, cropping, blurring, contrasting, and flipping, which increased the model's accuracy by 6.5\%. They also employed weight regularization, batch normalization, and dropout to optimize performance.

Addressing the difficulty of transferring models trained on one type of cell (monocrystalline) to another (polycrystalline), Xie et al.\cite{xie2023effective} employed a CNN architecture based on ResNet-50, modifying its last fully connected layer for their specific task. They introduced an attention-based transfer learning approach and class-aware domain discriminators to enhance the model's ability to generalize across domains. This method effectively improved performance on polycrystalline cells, achieving an F1-score of 0.8734, a recall of 84.7\%, and a precision of 90.15\%. Data augmentation techniques were also applied to enhance robustness. However, reliance on pre-trained models like ResNet-50, originally optimized for general image classification tasks, might constrain performance for this specialized application. Additionally, the integration of attention mechanisms and adversarial learning increases computational complexity.

Similarly, Chen et al.\cite{chen2020solar} also proposed a multi-spectral CNN for detecting surface defects in polycrystalline silicon solar cells, which present challenges due to their heterogeneous texture and complex background. They optimized the CNN by adjusting depth and convolution kernel sizes to effectively identify defect features. The model achieved an overall accuracy of 94.30\%, with significantly higher precision and recall compared to traditional methods. However, it exhibited lower detection rates for small and linear defects like broken gates and scratches, indicating weaker feature extraction for these types.

Chindarkkar et al.\cite{chindarkkar2020deep} addressed the challenge of adapting models trained on high-resolution lab EL images to perform well on lower-resolution field images. They applied a deep learning-based approach using the ResNet-50 model to detect defects in solar cells under field conditions, achieving an impressive accuracy of 98.59\% when trained and tested with field data. Data augmentation was applied to address data imbalance, increasing the dataset size eightfold. However, the high resource demands of the ResNet architecture render it for deployment in manufacturing environments.

 While exploring the use of CNNs and SVMs for automating the classification of defects in PV cells using EL images, the CNN model proposed by Ahmad et al.\cite{ahmad2020photovoltaic} achieved an accuracy of 91.58\%. Their model significantly outperformed the SVM models. The architecture consisted of nine convolutional layers with increasing numbers of kernels, followed by five max-pooling layers and a fully connected layer with 1,024 neurons. While effective, the addition of more convolutional kernels considerably increases computational demands.

\begin{table}[t]
\caption{Study of various CNN Architectures for PV Cell Defect Detection}
\label{tab:comparative_study}
\vspace{3mm} 
\centering
\scriptsize
\begin{tabular}{|p{2cm}|p{4cm}|p{1.5cm}|p{3cm}|p{1.2cm}|p{3cm}|}
\hline
\textbf{Authors} & \textbf{Contribution} & \textbf{Data Size} & \textbf{PV Cell Texture} & \textbf{Accuracy} & \textbf{Architecture} \\
\hline
Xie et al.\cite{xie2023effective} & Improvement of defect detection model by employing transfer learning with adversarial and attention-based mechanisms to adapt a model across different cell types & 2624 & EL images consisting of monocrystalline and polycrystalline silicon solar cells & 90.15\% & ResNet-50 with 2048-neuron input and a 2-neuron output \\
Acharya et al.\cite{acharya2021deep} & A Siamese CNN to classify various types of defects such as no defect, micro crack, large-scale defects, and low-resolution defects & 2624 & Monocrystalline and polycrystalline EL images & 74.75\% & Siamese CNN consists of five convolutional layers and a fully connected layer \\
Akram et al.\cite{akram2019cnn} & Light CNN architecture for automatic PV defect detection & 2624 & Monocrystalline and polycrystalline EL images & 93.02\% & Consisted of 6 layers, 4 convolutional layers and fully connected layers \\
Chen et al.\cite{chen2020solar} & Multispectral CNN for detection of various defects under different spectra & 15,330 non-defective images and 5,915 defective images & Polycrystalline cells & 94.3\% & Multispectral CNN with multiple convolutional layers with different kernel sizes (ranging from 3x3 to 7x7) and three parallel feature extraction branches \\
Chindarkkar et al.\cite{chindarkkar2020deep} & Deep learning-based model for detecting cracks on PV cells in low-resolution images implying real-life condition & 8768 & Low-resolution field images of PV cells from different manufacturers covering different climatic zones & 98.59\% & ResNet-50 \\
Ahmad et al.\cite{ahmad2020photovoltaic} & Enhances power system efficiency by automating defect identification by employing CNN and SVM targeting seven classes of defects & 2624 & Monocrystalline and polycrystalline EL images & 91.58\% & Nine-layered CNN architecture \\
Zahid et al.\cite{zahid2023lightweight} & A custom efficient lightweight CNN architecture to detect micro-cracks and other defects in PV cells that is deployable within PV manufacturing facilities & 294, 1584 (after augmentation) & - & 96\% & Custom CNN with 2-convolutional layers and 2 fully connected layers \\
Fan et al. \cite{fan2022automatic} & Introduces a feature fusion model that aggregates low-level features and deep semantically strong features using a self-attention mechanism to enhance the accuracy of micro-crack detection & 10970 & Polycrystalline cells & 99.11\% & ResNet-50 \\
Hussain, Al-Aqrabi et al. \cite{hussain2022pv} & PV-CrackNet, Custon CNN to detect defects in PV cell and a filter-induced augmentation method as a cost-effective solution for creating representative data samples & 787 & Polycrystalline Cells with different configuration & 97\% & Custom CNN with only 2 Convolutional layer \\
\hline
\end{tabular}
\end{table}

Then focusing on computational efficiency, Zahid et al.\cite{zahid2023lightweight} presented a custom CNN with two convolutional blocks followed by two fully connected layers. Designed using a bottom-up approach, the final model contained 8.76 million parameters—significantly fewer than state-of-the-art models like VGG-19, which has over 143 million parameters. Incorporating batch normalization resulted in the highest performance, achieving a precision of 96\%, a recall of 100\%, and an F1 score of 98\%.

Hassan and Dhimish \cite{hassan2023dual} on the contrary, introduced the Dual Spin Max Pooling CNN (DSMP-CNN) to improve the detection of defects such as cracks, microcracks, and Potential Induced Degradation (PID) in PV cells. By leveraging a dual-spin max pooling mechanism, the model captured more intricate features in EL images, achieving a validation accuracy of 96.97\%. Despite its success, limitations include the need for a more diverse dataset to ensure generalizability.

Zhang and Yin \cite{zhang2022solar} addressed data scarcity and model complexity by combining data augmentation using a Deep Convolutional Generative Adversarial Network (DCGAN) with a lightweight CNN model based on a modified VGG16 architecture. They replaced traditional convolutions with depthwise separable convolutions to reduce the number of parameters. This approach significantly improved the model's accuracy from 47\% to approximately 77\% and decreased testing time from 57 milliseconds to 22 milliseconds per image. Zhang et al.\cite{zhang2020micro} also investigated the use of transfer learning for detecting micro-cracks in polycrystalline solar cells. Utilizing a Deep Adaptation Network (DAN) model based on ResNet-50, they adapted a model pre-trained on monocrystalline cells to the more complex polycrystalline cells. Data augmentation addressed class imbalances, and the model achieved an average accuracy of 76.2\%, outperforming other methods.

To address data scarcity, Hussain, Al-Aqrabi et al. \cite{hussain2022pv} introduced a filter-induced augmentation method as a cost-effective solution for creating representative data samples. Their custom architecture was computationally efficient, with just 7.01 million trainable parameters, yet delivered an outstanding F1 score of 97\%. Another study by Hussain et al.\cite{hussain2022gradient} proposed a novel CNN architecture designed to identify microcracks on PV cell surfaces, attaining an impressive F1 score of 98.8\%. The architecture outperformed state-of-the-art models in most evaluated criteria, including computational efficiency and post-deployment performance.

In their study, Fan et al.\cite{fan2022automatic} provided a detailed analysis of automatic micro-crack detection in polycrystalline solar cells in industrial settings. Using a ResNet-based architecture with a self-attention mechanism, they achieved 99.11\% accuracy, significantly outperforming traditional deep neural network approaches. The method effectively captured precise geometric details crucial for defect detection.

In a comprehensive review, Hussain et al.\cite{hussain2023review} highlighted the shift towards using CNN architectures for automating defect detection in EL-based PV cells, emphasizing their advantages like automated feature extraction, improved generalization, reduced labor costs, and increased accuracy. It stresses the need for benchmark PV datasets to validate models reliably, ensuring true generalization across diverse data. While many studies with 99\% accuracy using only 777 images, show promising results, concerns remain about generalization to broader data distributions. The review suggests focusing on custom CNNs that are both high-performing and computationally efficient for easier deployment on standard hardware. Additionally, recent research compares new architectures against existing models on metrics like computational complexity to ensure practical application.

In summary, CNNs have demonstrated considerable potential in detecting defects in PV cells, with approaches ranging from complex architectures like Deep Siamese CNNs \cite{acharya2021deep} to efficient lightweight models \cite{hussain2022pv} suitable for real-time applications. Innovations such as multi-spectral CNNs \cite{chen2020solar}, transfer learning \cite{xie2023effective}, and domain adaptation have enhanced adaptability and accuracy under different conditions. Table \ref{tab:comparative_study} presents an overview of the use of CNN to detect defects in PV cells in different literature. Challenges remain in generalizing across various cell types and achieving computational efficiency, as many models are resource-intensive and unsuitable for real-time industrial use. Additionally, the scarcity of comprehensive datasets also limits effective training \cite{hussain2023review, hassan2023survey}, hindering application in diverse manufacturing scenarios. Therefore, future research should focus on developing computationally efficient models capable of high performance with limited data resources to meet practical production demands.

\section{Paper Contribution}
The research focuses on developing a computationally efficient CNN architecture designed for defect detection in PV cells. This model is optimized for deployment on resource-constrained edge devices, which are common in industrial production facilities. 
The study also handles the challenge of data scarcity by implementing advanced data augmentation techniques that represent the deployment scenario encountered in the production facility, increasing the dataset from 228 to 538 samples. This shows how data augmentation can be leveraged to achieve better performance as collecting real data from PV production could be tough and require expertise. The proposed architecture proved to be lightweight, containing 2.92 million learnable parameters, ensuring rapid processing while maintaining high defect detection accuracy. The model was tested and achieved a precision of 91\%, recall of 89\%, and an F1-score of 90\%, showing strong performance in identifying micro-cracks and contaminations in PV cells.

\section{Methodology}
This section will detail the development and implementation of PV-faultNet CNN architecture designed for efficient defect detection in PV cells. We will describe the model's architecture, including the number and types of layers, and explain how we optimized it for deployment on resource-limited production devices. The section will also cover our data preparation process, including the comprehensive data augmentation techniques used to address data scarcity and enhance model generalization. We'll discuss the training process, including the choice of hyperparameters, optimization algorithms, and any regularization techniques employed. Finally, we'll outline our evaluation metrics and the experimental setup used to assess the model's performance in terms of precision, recall, and F1 score.

\subsection{Dataset}

Table \ref{table:original_data} represents the original dataset consisting of two classes: defective and normal.

\begin{table}[h!]
\centering
\caption{ Original dataset. }
\vspace{3mm} 
\begin{tabular}{|c|c|}
\hline
\textbf{Class} & \textbf{Sample Count} \\
\hline
Defective & 153 \\
Normal & 75 \\
\hline
\end{tabular}
\label{table:original_data}
\end{table}

Using the existing dataset, it can be concluded that it is not sufficient to utilize it for the development of a highly generalized and reliable model, that would be able to distinguish between the classes.

\begin{figure}[h]
    \centering
    \includegraphics[width=0.5\linewidth]{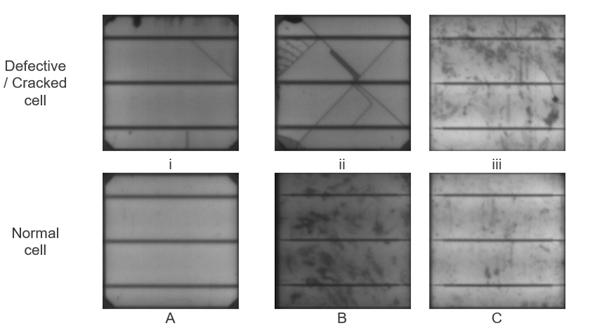}
    \caption{Examples of PV cells from Original Dataset}
    \label{fig:original_data}
\end{figure}

Figure \ref{fig:original_data} presents a sample set of PV cell surfaces, showing both defective and normal cells. Before adjusting the original dataset for scaling, it was crucial to understand the visual differences between classes and the variation within each class. The images reveal surface heterogeneity, particularly in texture. For instance, normal cells (A) and (C) display clearer surface details than the middle image (B), highlighting texture differences that could mislead the model into associating clarity with the normal class.

Comparison between defective and normal cells in Figure \ref{fig:original_data} shows that internal and external factors, such as shading or filter quality, significantly affect the image. For example, shading in a normal cell may resemble micro-cracks in defective cells, potentially causing misclassification. Additionally, it emphasizes the importance of the busbar, which directs energy flow on the cell surface but varies in style. Normal cell (A) has a 'cross-surface' busbar, while normal cell (C) has a 'cut-off' style. Variations in busbar appearance, such as lighter or darker tones, can also lead to confusion, as lighter busbars might be mistaken for cracks, suggesting defects.

\subsection{Data Augmentation}
A thorough data analysis suggested that the variability within the dataset, both globally and within classes, could be better addressed through representative data modeling rather than simply expanding the dataset. Consequently, the dataset was increased to 538 samples from the original 228 images through targeted augmentations, reflecting the practical challenges of accessing PV data from manufacturing facilities and the scarcity of open-source data. The data augmentations fell into two main categories: translational invariance and translational equivariance.

Translational invariance is represented by Hussain et. al. \cite{hussain2022pv} as denoted in equation \ref{eq:1}:

\begin{equation}
I(t(m)) = I(m)
\label{eq:1}
\end{equation}

where $I$ represents the function for the image $m$, and $t$ represents the transformation which was employed on it. 
The architecture's internal layer design relied solely on translational invariance, which did not alter the dataset physically but effectively expanded it. This approach preserved regional transformations through aggregation, making it beneficial during the design phase as information moves through deeper architectural layers. Conversely, translational equivariance was selected as the framework for data scaling which is depicted in equation \ref{eq:2}:

\begin{equation}
I(t(m)) = t(I(m))
\label{eq:2}
\end{equation}

where $I$ signifies the function for the image $m$, and $t$ signifies the employed transformation. When it comes to translational equivariance, it's clear that the input image will change depending on the specific transformation applied, like the equation for translational invariance.

Each type of augmentation was chosen because it's likely to happen in a PV manufacturing complex, taking into account factors like different production line setups and EL camera specs.

\subsubsection{Augmentation Setup}
To apply data augmentation on the original dataset, Roboflow \cite{Roboflow2024} was used. Roboflow is a user-friendly online platform that facilitates data handling, pre-processing, augmentation, and versioning. The platform supports various data formats and offers tools for resizing, normalization, format conversion, and applying augmentation techniques. Additionally, Roboflow provides dataset versioning and seamless export capabilities to machine learning frameworks.

\subsubsection{Augmentation Application}
\begin{figure}[h]
    \centering
    \includegraphics[width=0.5\linewidth]{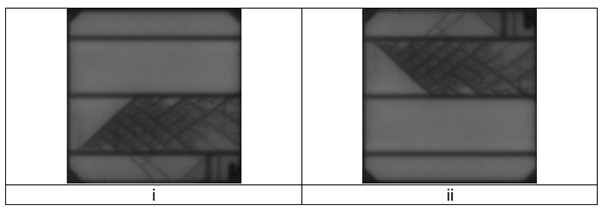}
    \caption{Geometric transformation: Vertical Flipping (i) before (ii) after}
    \label{fig:flipping}
\end{figure}

PV cells go through multiple stages, from silicon ingot formation to wafer slicing and various treatments. It can create images with different orientations. To capture this variation, vertical flipping was used to generate representative instances, as shown in Figure \ref{fig:flipping}.

\begin{figure}[h]
    \centering
    \includegraphics[width=0.5\linewidth]{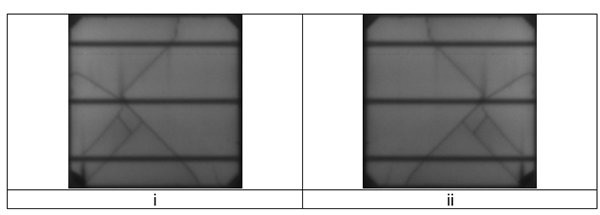}
    \caption{Geometric transformation: Horizontal Flipping (i) before (ii) after}
    \label{fig:horizontal_flip}
\end{figure}

Figure \ref{fig:horizontal_flip} demonstrates the use of horizontal orientation, similar to the rationale for vertical flipping, considering factors like production line constraints and EL camera setup. These adjustments were applied selectively to avoid unnecessary repetition.

Additional augmentations, such as random brightness (-25\% to +25\%), exposure (-15\% to +15\%), Gaussian blur (0 to 3.5 pixels), and salt and pepper noise (1.8\% of pixels), were designed to replicate real-world conditions in PV manufacturing. These simulate variations in lighting, camera settings, minor equipment misalignments, and random pixel disturbances like dust or electronic interference. These techniques help create a more robust dataset, enabling the model to perform well under diverse and imperfect conditions during PV cell inspections.

\begin{table}[h!]
\centering
\caption{ Augmented dataset. }
\vspace{3mm}
\begin{tabular}{|c|c|}
\hline
\textbf{Class} & \textbf{Sample Count} \\
\hline
Defective & 361 \\

Normal & 177 \\
\hline
\end{tabular}
\label{table:Augmented dataset}
\end{table}

Table \ref{table:Augmented dataset} illustrates the distribution of the modified dataset in terms of the number of samples within each category. The dataset comprised a combined total of 538 samples post-augmentation increasing the dataset.

\subsection{Proposed Architecture}
The study proposes a CNN model suitable for deployment on resource-constrained edge devices used in production facilities, while also mitigating the challenge of limited relevant data in this domain. The specific objectives include designing a CNN architecture optimized for production devices, addressing data scarcity through effective data augmentation techniques, and assessing the model's performance in terms of accuracy and computational complexity. It focused on crafting a unique CNN structure instead of taking the easy road with transfer learning, aiming to simplify things by using fewer convolutional blocks and fine-tuned filters.  

 The training of the model was conducted on Google Collaboratory \cite{GoogleColab2024}, chosen for its free GPU access, though time constraints limited the maximum number of epochs to 50, with a learning rate set at 0.02. The hyper-parameters were established at a global level to ensure an impartial performance evaluation are detailed in Table \ref{tab:global_hyperparameters}.

\begin{table}[h!]
\centering
\caption{Global Hyper-parameters}
\vspace{3mm}
\begin{tabular}{|c|c|}
\hline
\textbf{Hyper-parameter} & \textbf{Value} \\
\hline
Batch size & 32 \\
Learning Rate & 0.02 \\
Optimizer & SGD-M \\
Decay Rate & 0.01 \\
\hline
\end{tabular}
\label{tab:global_hyperparameters}
\end{table}

\begin{figure}[h]
    \centering
    \includegraphics[width=0.8\linewidth]{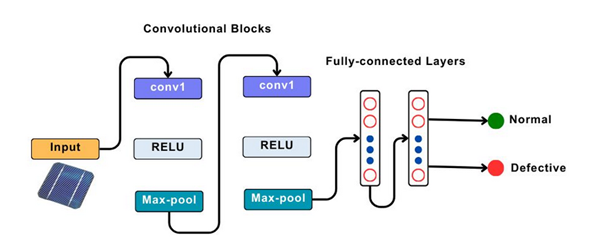}
    \caption{Proposed CNN Architecture.}
    \label{fig:architecture}
\end{figure}

The overall view of the proposed CNN architecture is shown in Figure \ref{fig:architecture} featuring two convolutional blocks and two fully connected layers. This was based on the results of the data examination, as previously discussed. The examination revealed that although there were differences at the overall and internal class levels, these differences mainly revolved around surface textures, light strengths, hardware influences, and busbar arrangements. As a result, using fewer convolutional blocks with carefully adjusted filters seemed like a good way to capture the essential unique features of the two classes.

 Based on the internal structure presented in Table \ref{tab:neural_net}, the suggested design consists of two convolutional blocks and two fully connected layers, parameters totalling around 2.92 million.
 
\begin{table}[h]
\centering
\caption{Internal block-wise architecture of the model}
\vspace{3mm}
\begin{tabular}{|c|c|c|}
\hline
\textbf{Layer} & \textbf{Output Dimensions} & \textbf{Total Learnable parameters} \\ \hline
Input          & 3,300x300                  & ---                                 \\
Convolution-01 & 5,298x298                  & 140                                 \\
Max-pool       & 5,149x149                  & ---                                 \\ 
Convolution-02 & 10,147x147                 & 460                                 \\ 
Max-pool       & 10,73x73                   & ---                                 \\ 
FC-01          & 100 neurons                & 2916100                             \\ 
ReLU           & ---                        & ---                                 \\ 
FC-02          & 50 neurons                 & 5050                                \\ 
ReLU           & ---                        & ---                                 \\ 
Output         & 2 neurons                  & 102                                 \\ \hline
\multicolumn{2}{|c|}{\textbf{Total Learnable parameters}} & 2.92 million \\ \hline
\end{tabular}
\label{tab:neural_net}
\end{table}

The initial layer had 5 filters, each consisting of 3 × 3 pixels. The decision to use odd-sized filters was deliberate, as it ensures a central pixel for encoding the filter output, unlike even-sized filters (e.g., 2 × 2 pixels) which can introduce aliasing errors due to the lack of a central pixel. The dimensions of the resulting feature maps from this first convolutional block were determined using a specific formula shown in equation \ref{eqn:3}:
\begin{equation}
  f_{\text{out}} = \left[ \frac{f_{\text{in}} - 2p - k}{s} \right] + 1
  \label{eqn:3}
\end{equation}
where, $f_{\text{out}}$ is output features, $f_{\text{in}}$ is input features, $p$ is padding dimension, $k$ is kernel and $s$ is stride.

The output of the first calculation served as the stepping stone for the second convolutional block, which amplified the number of filters. This allowed for deeper feature extraction, with the first block targetted on basic feature detection like lines and edges, and the second block polishing these features for the two fully connected layers.
Max-pooling was incorporated to achieve translational invariance by aggregating local features and reducing positional dependencies. A stride of 2 was used in the max-pooling operation, leading to a 50\% reduction in the spatial dimensions of the output, as indicated by the equation. ReLU was selected as the activation function for its advantages over Sigmoid and TanH in mitigating the vanishing gradient problem.
The number of learnable parameters for the convolutional layers was determined using the formula denoted in equation \ref{eqn:4}:
\begin{equation}
  (p \times q \times f + 1) \times k
  \label{eqn:4}
\end{equation}
where $p$ and $q$ are the filter sizes, $f$ is the number of input features, and $k$ is the number of output feature maps. 

Following the second convolutional block, the output was flattened before being passed into the fully connected layers. Therefore, the filter dimensions were no longer relevant in calculating the learnable parameters for these layers, which were computed following equation \ref{eqn:5}:
\begin{equation}
  (N_i + 1) \times N_o
  \label{eqn:5}
\end{equation}
where $N_i$ refers to the input neurons and $N_o$ to the output neurons.

\subsection{Epochs}
This section presents an analysis of the custom CNN model's performance on the augmented dataset using a bottom-up approach, progressively increasing the number of epochs to optimize model accuracy and generalization.

\begin{figure}[h]
    \centering
    \begin{minipage}{0.45\textwidth}
        \centering
        \caption{Epoch 5: Confusion Matrix}
        \includegraphics[width=\linewidth]{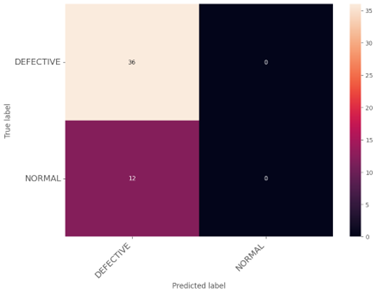}
        \label{fig:epoch-5-CF}
    \end{minipage}
    \hfill
    \begin{minipage}{0.45\textwidth}
        \centering
        \captionof{table}{Epoch 5: Performance Metrics}
        \vspace{3mm}
        \begin{tabular}{|c|c|}
            \hline
            \textbf{Metric} & \textbf{Value} \\
            \hline
            Precision & 75\% \\
            Recall & 100\% \\
            F1-Score & 86\% \\
            Train Accuracy & 67.76\% \\
            Valid Accuracy & 79.17\% \\
            \hline
        \end{tabular}
        \label{tab:performance_epoch5}
    \end{minipage}
\end{figure}

\textbf{Epoch 5:} The initial assessment presented in Table \ref{tab:performance_epoch5} revealed that training accuracy for epoch 5 stabilized around 67.79\%, while validation accuracy peaked at 79\%, suggesting good generalization but under-fitting. The confusion matrix in Figure \ref{fig:epoch-5-CF} indicated a bias toward predicting defective samples, with no correct classifications of normal instances. The precision was 75\%, recall 100\%, and F1-score 86\%, indicating a need for further adjustments to address the class imbalance and improve normal instance detection.

\begin{figure}[h]
    \centering
    \begin{minipage}{0.45\textwidth}
        \centering
        \caption{Epoch 15: Confusion Matrix}
        \includegraphics[width=\linewidth]{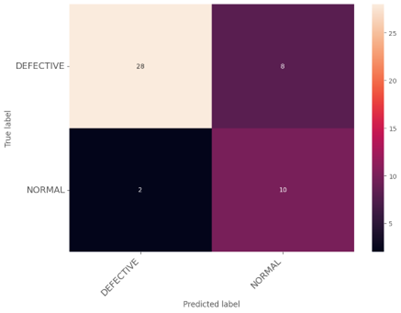}
        \label{fig:epoch-15-CF}
    \end{minipage}
    \hfill
    \begin{minipage}{0.45\textwidth}
        \centering
        \captionof{table}{Epoch 15: Performance Metrics}
        \vspace{3mm}
        \begin{tabular}{|c|c|}
            \hline
            \textbf{Metric} & \textbf{Value} \\
            \hline
            Precision & 93\% \\
            Recall & 78\% \\
            F1-Score & 85\% \\
            Train Accuracy & 89.38\% \\
            Valid Accuracy & 89.58\% \\
            \hline
        \end{tabular}
        \label{tab:performance_epoch15}
    \end{minipage}
\end{figure}

\textbf{Epoch 15:} Performance metrics in Table \ref{tab:performance_epoch15} showed some improvement, with training and validation accuracy closely aligned at approximately 89\%, suggesting reduced over-fitting. The precision increased to 93\%, while the recall dropped to 78\%, further validating the confusion matrix, as in Figure \ref{fig:epoch-15-CF}, which shows some defective cases were missed. The F1-score of 85\% highlighted the need for enhanced recall to ensure better detection of defective instances.

\begin{figure}[h]
    \centering
    \begin{minipage}{0.45\textwidth}
        \centering
        \caption{Epoch 35: Confusion Matrix}
        \includegraphics[width=\linewidth]{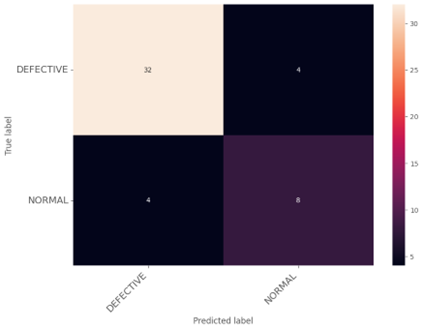}
        \label{fig:CMepoch35}
    \end{minipage}
    \hfill
    \begin{minipage}{0.45\textwidth}
        \centering
        \captionof{table}{Epoch 35: Performance Metrics}
        \vspace{3mm}
        \begin{tabular}{|c|c|}
            \hline
            \textbf{Metric} & \textbf{Value} \\
            \hline
            Precision & 89\% \\
            Recall & 89\% \\
            F1-Score & 89\% \\
            Train Accuracy & 99.79\% \\
            Valid Accuracy & 89.58\% \\
            \hline
        \end{tabular}
        \label{tab:performance_filter5_10_epoch35}
    \end{minipage}
\end{figure}

\textbf{Epoch 35: }At epoch 35, the training accuracy reached 99.79\%, while validation accuracy remained at 89.58\%, indicating potential over-fitting, as shown in Table \ref{tab:performance_filter5_10_epoch35}. Both precision and recall balanced at 89\%, reflecting a more stable model performance, though some miss-classifications persisted as per Figure \ref{fig:CMepoch35}. The loss and accuracy results suggested a need for further regularization to improve generalization.

\begin{figure}[h]
    \centering
    \begin{minipage}{0.45\textwidth}
        \centering
        \caption{Epoch 50: Confusion Matrix}
        \includegraphics[width=\linewidth]{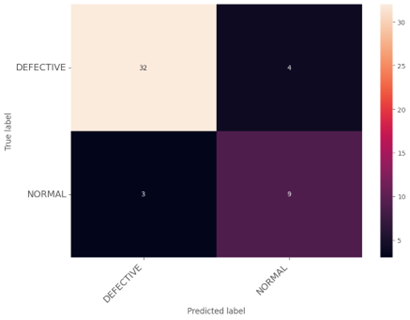}
        \label{fig:CMepoch50}
    \end{minipage}
    \hfill
    \begin{minipage}{0.45\textwidth}
        \centering
        \captionof{table}{Epoch 50: Performance Metrics}
        \vspace{3mm}
        \begin{tabular}{|c|c|}
            \hline
            \textbf{Metric} & \textbf{Value} \\
            \hline
            Precision & 91\% \\
            Recall & 89\% \\
            F1-Score & 90\% \\
            Train Accuracy & 100\% \\
            Valid Accuracy & 91.67\% \\
            \hline
        \end{tabular}
        \label{tab:performance_filter5_10_epoch50}
    \end{minipage}
\end{figure}

\textbf{Epoch 50: }As depicted in Table \ref{tab:performance_filter5_10_epoch50}, the model achieved a precision of 91\%, recall of 89\%, and an F1-score of 90\%. Despite reaching a perfect training accuracy of 1.0, the validation accuracy was slightly lower at 91.67\%, reinforcing concerns about over-fitting. Adjustments such as batch normalization improved recall to 94\%, indicating better detection of defective cases, yet additional fine-tuning would have been beneficial for optimal generalization. The model's confusion matrix is shown in Figure \ref{fig:CMepoch50}.

\begin{figure}[h]
    \centering
    \includegraphics[width=0.85\linewidth]{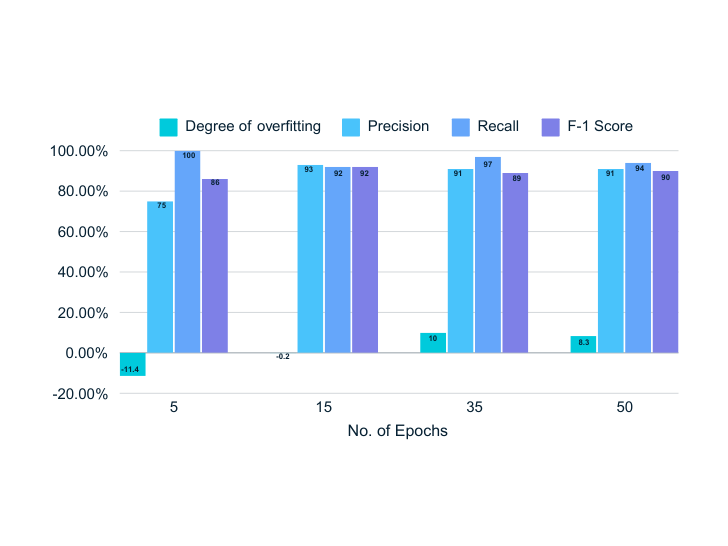}
    \caption{Different evaluation matrices across different epochs}
    \label{fig:chart}
\end{figure}

After analyzing the model's performance at different epochs, the results are summarized in Figure \ref{fig:chart} to provide an overall view. At epoch 5, high recall but low precision indicated many false positives, with significant over-fitting. By epoch 15, precision and recall balanced around 92\%, with a notable reduction in over-fitting, showing better generalization. At epoch 35, recall reached 97\%, though a slight rise in over-fitting suggested a preference for training data. By epoch 50, precision and recall were balanced at around 91\% and 89\%, with moderate over-fitting, showing refined model performance. The chart highlights the trade-off between model accuracy and over-fitting as training progressed.

\section{Results}
At epoch 50, the model performed in a balanced and efficient manner achieving precision and recall of 91\% and 89\% with an F1-score of 90\%. 
The precision indicates that 90\% of the defect predictions were correct which is essential as it helps to minimize false positive or misclassification of normal cells. Recall suggests that the model successfully identified 89\% of all actual defect. F1-score of 90\% highlights the model’s balanced performance, suggesting it is adept at accurately detecting defects without a strong bias toward either precision or recall. The F1-score is particularly useful in situations like PV defect detection, where both missed detections (due to low recall) and false positives (due to low precision) have significant operational impacts. Furthermore, The model reached 100\% training accuracy, indicating that it fit the training data perfectly. However, the validation accuracy was 91.67\%, a slight decline from the training accuracy, which signals some degree of over-fitting.  
Batch normalization was applied to observe if it enhanced the result where there was a visible decline in precision and F1-score but a slight increase of 3\% in the recall. Also, the validation accuracy reduced significantly from 91.67\% to 87\% keeping the training accuracy at 100\% decreasing the overall generalisability. Furthermore, 25\% dropout was applied to disable random neurons to enable better generalization. Additionally, batch normalization coupled with dropout was applied to observe the impact. However, no notable performance improvement was observed. These tests confirmed that the original model without any hyper-parameter tuning was the most efficient setup.

To explore the impact of data scarcity on model performance, the original dataset was trained on 50 epochs. This assessment aimed to demonstrate how limited data affects the ability of the model to generalize. With the performance of the original dataset presented in Table \ref{tab:original_data_performance_at_epoch50}, the model achieved a training accuracy of 100\%  and a validation accuracy of 84\%, alongside a precision of 69\%, recall of 86\%, and an F1-score of 77\%. These results highlight a clear tendency toward over-fitting, where the model performed perfectly on training data but struggled to maintain similar accuracy on unseen validation data. 

\begin{table}[h]
\centering
\caption{Performance of Original dataset at Epoch 50}
\vspace{3mm}
\begin{tabular}{|c|c|}
\hline
\textbf{Metric} & \textbf{Value} \\
\hline
Precision & 69\% \\
Recall & 86\% \\
F1-Score & 77\% \\
Train Accuracy & 100\% \\
Valid Accuracy & 84\% \\
\hline
\end{tabular}
\label{tab:original_data_performance_at_epoch50}
\end{table}

\begin{figure}[h]
    \centering
    \includegraphics[width=0.5\linewidth]{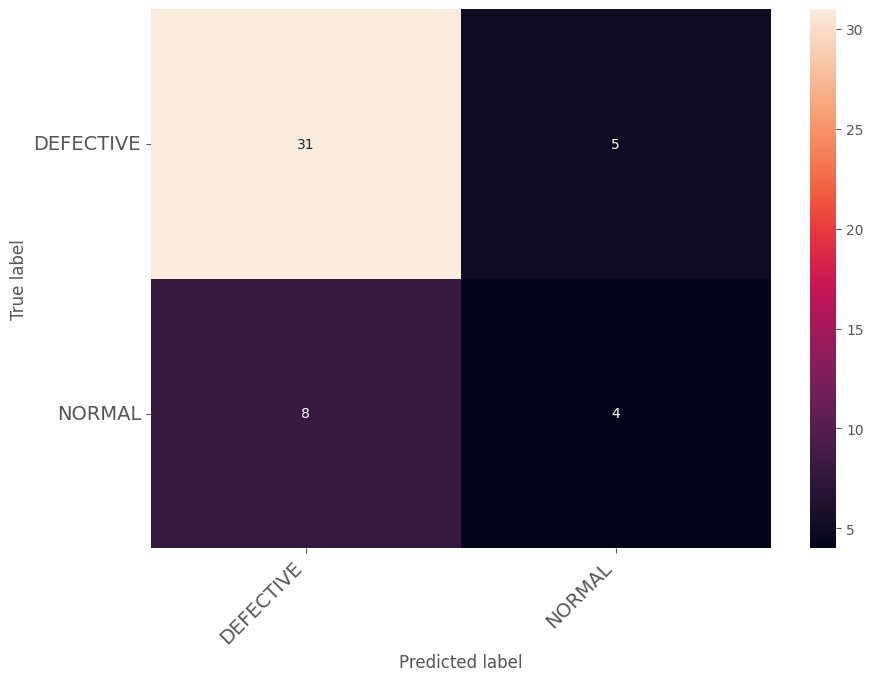}
    \caption{Confusion Matrix of Original Dataset}
    \label{fig:CMoriginal}
\end{figure}

The confusion matrix in Figure \ref{fig:CMoriginal} further illustrates the model's bias toward predicting the defective class, as it tends to misclassify normal instances more frequently than defective ones. Despite a relatively high recall for defective cases, the lower precision for the normal class indicates challenges in accurately distinguishing between the two categories. This tendency to over-predict defects could be detrimental in a production environment, leading to unnecessary rejections of non-defective cells, potentially increasing costs and reducing efficiency.

The comparison with the augmented dataset shows that data augmentation and regularization effectively mitigate data scarcity, improving model generalization. A richer dataset led to more balanced precision and recall, highlighting the importance of augmentation in addressing limited training data. This approach enhances the model's reliability for real-time defect detection in PV cell manufacturing.

\section{Discussion}
The analysis section of this study emphasized the effectiveness of the proposed lightweight CNN architecture, designed specifically for detecting defective PV cells with computational efficiency in mind. Unlike traditional models that prioritize only high accuracy, this study aimed to develop a streamlined architecture suitable for deployment on resource-constrained edge devices. To evaluate its efficiency, the architecture was compared to several benchmark models, focusing on the computational complexity quantified by the number of learnable parameters, as summarized in Table \ref{tab:computational_efficiency}.

\begin{table}[h!]
\centering
\caption{Comparison of Computational Efficiency}
\vspace{3mm}
\begin{tabular}{|c|c|}
\hline
\textbf{Architecture} & \textbf{Parameters (in million)} \\
\hline
ResNet50 \cite{reddy2019transfer} & 23.58 \\
VGG16 \cite{swasono2019classification}& 138.35 \\
GoogleNet \cite{gao2021blnn} & 6.8 \\
PV-CrackNet \cite{hussain2022pv}& 7.01 \\
PV-faultNet & 2.92 \\
\hline
\end{tabular}
\label{tab:computational_efficiency}
\end{table}

The comparison reveals that the proposed custom CNN model has the lowest number of parameters at 2.92 million, making it significantly lighter than other architectures like VGG-16 \cite{swasono2019classification}, which has 138.35 million parameters. This reduction in parameters directly impacts the computational resources required, shortening training time and lowering the hardware requirements. ResNet50\cite{reddy2019transfer} followed with 23.58 million parameters, while PV-CrackNet\cite{hussain2022pv}, despite having only two convolutional blocks and two fully connected layers, still required 7.01 million parameters due to its specific filter configuration.
This outcome underscores the advantage of the proposed custom CNN in balancing simplicity and efficiency, enabling real-time inspection capabilities on edge devices for scalable and effective quality control in PV cell manufacturing.

\section{Conclusion}
This study introduces a computationally efficient CNN architecture tailored for defect detection in solar PV cells, optimized for deployment on resource-limited edge devices. The model achieves a competitive balance between accuracy and efficiency, with only 2.92 million parameters along with precision of 91\%, recall of 89\% and F1-score 90\%, significantly outperforming larger models like VGG-16\cite{swasono2019classification} and ResNet50\cite{reddy2019transfer} in terms of computational demand. This enables real-time defect detection, crucial for scalable quality control in PV manufacturing.

The research emphasizes the importance of data augmentation in mitigating the challenges of data scarcity ~\cite{hussain2023child}. Comparisons between the original and augmented datasets demonstrated that targeted augmentation improves model generalization and balance between precision and recall, reducing over-fitting and enhancing detection accuracy. It has significant implications for industrial applications, offering a lightweight yet effective solution for automated quality inspection in PV manufacturing\cite{hussain2024depth}. Future work should focus on further optimizing the architecture, expanding datasets, and validating the model's performance in real-world production environments to enhance robustness and scalability.

\vspace{6pt} 


\bibliographystyle{unsrt}  
\bibliography{ref}  

\end{document}